# Greedy Algorithms for Approximating the Diameter of Machine Learning Datasets in Multidimensional Euclidean Space

Ahmad B. A. Hassanat
IT Department, Mutah University, Mutah – Karak, Jordan, 61710
ahmad.hassanat@gmail.com

Abstract

Finding the diameter of a dataset in multidimensional Euclidean space is a well-established problem, with well-known algorithms. However, most of the algorithms found in the literature do not scale well with large values of data dimension, so the time complexity grows exponentially in most cases, which makes these algorithms impractical. Therefore, we implemented 4 simple greedy algorithms to be used for approximating the diameter of a multidimensional dataset; these are based on minimum/maximum $l_2$ norms, hill climbing search, Tabu search and Beam search approaches, respectively. The time complexity of the implemented algorithms is near-linear, as they scale near-linearly with data size and its dimensions. The results of the experiments (conducted on different machine learning data sets) prove the efficiency of the implemented algorithms and can therefore be recommended for finding the diameter to be used by different machine learning applications when needed.

## 1. Introduction

The K-nearer neighbor (KNN) classifier, its variants and the nearest neighbor search approaches in general are some of the most used approaches in machine learning for their simplicity, common use and various applications. These approaches depend mainly on finding similarities in feature space; Euclidean distance is one of the most used. Sometimes, finding similarity is the only option available, as in content-based image retrieval (Hassanat & Tarawneh, 2016).

Finding the diameter in the Euclidean feature space is vital for some methods in different machine learning tasks, such as clustering and classification. These applications include, but are not limited to, data clustering for images database (Gudivada & Raghavan, 1995), pattern recognition, web clustering (Broder, Glassman, Manasse, & Zweig, 1997), Outlier Detection (Aggarwal, 2016), approximate furthest neighbor applications (Pagh, Silvestri, Sivertsen, & Skala, 2015), Euclidean graphs (Supowit, 1990) and computational geometry (Williams, 2018).

The diameter or furthest pair problem can be defined as follows: given a finite set *S* of points of size n in d-dimensional Euclidean space $E^d$, find the maximum Euclidean distance between two pairs of points from data points given. This problem is different from the furthest neighbor problem (Pagh, Silvestri, Sivertsen, & Skala, 2017).

This problem is solvable using a brute force algorithm (BF), which is based on comparing the distance between each point and all the other points, and after all comparisons, it returns the maximum distance. The time complexity is therefore O(n.n.d), which is impractical for data with large values of n and/or d. On the contrary, the nearest-neighbors problem is solved optimally in quasilinear time O(n log n); however, an optimal solution in similar time for the diameter problem is not achieved so far (Agarwal, Matoušek, & Suri, 1992) and (Williams, 2018).

We found several efficient algorithms in the literature for the diameter problem if the data is of two dimensions such as the work of (Preparata & Shamos, 1985), or three dimensions such as the works of (Clarkson & Shor, 1989), (Chazelle, Edelsbrunner, Guibas, & Sharir, 1992), (Matoušek & Schwarzkopf, 1993), (Ramos E. , 1997), (Bespamyatnikh, 1998) and (Ramos E. A., 2000); nevertheless, they all cannot be extended to higher dimensional spaces. (Finocchiaro & Pellegrini, 2002).

Less efficient algorithms, regardless the size of the dimension, include the work of (Yao, 1982), who proposed several algorithms including the minimum spanning tree and the diameter in d dimension with $O(n^2)$ time. With approximation of $\sqrt{3}$ (Eğecioğlu & Kalantari, 1989) proposed an iterative approach costs O(n.d) for each iteration with $m$ iterations, where m ≤ n. They showed that $m$ might reach $n$ in the worst case and thus become $O(d.n^2)$. Yao's algorithm is faster than the BF, since the d is not counted here. While the performance of Eğecioğlu and Kalantari's algorithm converges to that of the BF in the worst case, when $m=n$, however, this is rare to happen and depends mainly on the dataset itself, as shown by (Eğecioğlu & Kalantari, 1989).

(Agarwal, Matoušek, & Suri, 1992), solved the problem with $(1 + \varepsilon)$-approximation in $O(n\ \varepsilon^{(1-k)/2} \log n)$ time. With a similar approximation, (Finocchiaro & Pellegrini, 2002) propose a solution with $O(d.n.\log n+n^2)$ time. In the same year, another similar approximation is achieved by (Chan, 2002) who proposed a recursive algorithm to solve the problem with $O(n+1/\varepsilon^{3(d-1)/2})$ time. The same approximation is also maintained by (Agarwal, Har-Peled, & Varadarajan, 2005) who utilize a paradigm called Coresets for approximating various extent measures of a set of points, this approximation is then used to find a number of different measures including the diameter, which can be calculated in $O(n+1/\varepsilon^{d-(3/2)})$ time. Similar approximation is also achieved by (Imanparast, Hashemi, & Mohades, 2016) with $O(n+1/\varepsilon^{d-2})$ time.

(Har-Peled, 2001) proposed a practical algorithm with quadratic time in worst case; however, the running time is sensitive to the input dataset and can be approximated by $O((n+1/\varepsilon^{3(d-1)/2}) \log 1/3)$.

According to (Williams, 2018), the best known algorithms for solving the furthest pair problem in d dimensional Euclidian space still have running time bounds of the form $O(n^{2-1/\theta(d)})$, which is barely sub-quadratic.

Most of the algorithms found in the literature are quadratic or sub-quadratic in terms of the number of points, or exponential in terms of the number of dimensions d, as these algorithms assume (in a way or another) that d << n. However, having large values of d (like in the case of machine learning datasets) makes such algorithms impractical, particularly in online applications. One of the reasons behind the long time consumed by these algorithms is due to the quality of approximation, as most of these algorithms attempt to satisfy a specific pre-defined approximation goal ε, where ε differs from paper to another, e.g. 0< ε ≤1 (Agarwal, Matoušek, & Suri, 1992), and √3 (Eğecioğlu & Kalantari, 1989), etc. Table 1 summarizes the time complexity of the well-known state-of-the-art algorithms.

The purpose of this paper is to trade off some quality by not being committed to a predefined constraint, implementing near-linear time approximate algorithms, which are based on greedy approaches to solve the diameter problem. The greedy approaches used are Minimum/Maximum

Norms, Hill climbing, Tabu search and Beam search. We need faster algorithms to be used for a larger project on approximate nearest neighbor classifier, and since such a classifier is approximate by its nature, a more approximate (lower quality) furthest pair algorithm may not affect its performance.

Table 1 Summary of time complexity of state-of-the-art diameter's algorithms of *n* points in *d* dimensions.

| Method | Reference | Time complexity | |
|---|---|---|---|
| M1 | (Imanparast, Hashemi, & Mohades, 2016) | $O(n + 1/\varepsilon^{d-2})$ | … (1) |
| M2 | (Agarwal, Matoušek, & Suri, 1992) | $O(n\varepsilon^{\frac{1-d}{2}} \log n)$ | … (2) |
| M3 | (Yao, 1982) | $O\left(n^{2-\alpha(d)} (\log n)^{1-\alpha(d)}\right)$ where $\alpha(d) = 2^{-(d+1)}$ | … (3) |
| M4 | (Eğecioğlu & Kalantari, 1989) | $O(nm)$ where $m \leq n$ | … (4) |
| M5 | (Finocchiaro & Pellegrini, 2002) | $O(dn \log n + n^2)$ | … (5) |
| M6 | (Agarwal, Har-Peled, & Varadarajan, 2005) | $O(n + 1/\varepsilon^{d-(\frac{3}{2})})$ | … (6) |
| M7 | (Chan, 2002) | $O(n + 1/\varepsilon^{3(d-1)/2})$ | … (7) |
| M8 | (Har-Peled, 2001) | $O\left(\left(n + \frac{1}{\varepsilon^{3(d-1)/2}}\right)\left(\log \frac{1}{\varepsilon}\right)\right)$ | … (8) |

## 2. Implementation of the greedy algorithms

The following algorithms are implemented using Microsoft Visual studio C++ and tested on several machine learning common datasets.

### *2.1 Minimum/Maximum $l_2$ norms.*

This algorithm is implemented based on finding the l2 norms for each point. Then it stores k points of those having the minimum l2 norms, and *k* points of those having the maximum l2 norms. Here, we choose *k* to be log n for time complexity purposes. Some machine learning datasets use negative values, and since the norm of a point with negative values might equal to a different point, we opt for translating all points by subtracting the minimum of each dimension of all points. The stored *k* points (log n) are then compared using brute force algorithm to find the furthest points (the diameter of the dataset). See Algorithm (1).

Finding the norms takes O(n.d) time, finding the *k* maximum and minimum norms takes O(2nlogn) time, and the brute force comparison O(d. log n . log n) time. The total time complexity of this algorithm is

$$T(n, d) = O(nd + n\log n + d(\log n)^2) \quad \ldots (9)$$

We cannot further simplify equation (9), because it depends on both n and d values, if *d* is a large value then the time complexity might be sub-linear, while if $n \gg d$ it might be a quasilinear time.

Algorithm 1: Minimum/Maximum $l_2$ norms (assuming all points are already shifted)
- - - - - - - - - - - - - - - - - - - - - - - - - - - - - - - - - - - - - - - - - - - - - - - - - - - - - - - - - - - - - - -
Input DATA: A data set of *n* points and *d* dimensions of real numbers, and a constant *k*.
Output MAX: The approximate diameter (the distance between the approximate furthest points)

---
1. For each point *i* in DATA create its $l_2$ $norm_i$
2. Store the points of the minimum k *norm* in $Set_1$
3. Store the points of the maximum k *norm* in $Set_2$
4. *MAX*=0
5. for each point *i* in $Set_1$
6 ….for each point *j* in $Set_2$
7…….*D*= $l_2$ ($Set_1[i]$, $Set_2[j]$)
8…….if *D* > *MAX* then *MAX=D*
9. Print *MAX*
---

## 2.2. Hill climbing implementation

This algorithm is inspired by the work of (Le Bourdais, 2015), which is about finding the furthest neighbors. This algorithm works by selecting a random point, then finding the furthest point by comparing all points, then the new point is taken to be compared with all the other points. Thus it keeps running that way keeping track of the points with the maximum distance until no further enhancement – i.e., no new points with a distance greater than the current distance. See Algorithm 2. This algorithm depends mainly on the number of iterations as each iteration costs O(n.d) time. The number of iterations depends on the nature of the data and the random starting point. The number of iterations is normally constant k << n, and therefore, the time complexity is

$$T(n,d) = O(k.n.d) \qquad \ldots (10)$$

Algorithm 2: Hill climbing
---
Input: DATA; A data set of *n* points and *d* dimensions of real numbers.
Output: MAX; The approximate diameter (the distance between the approximate furthest points)
---
1. $P_1$ = random point (1, n)
2. *MAX*=0
3. for each point $p_i$ in *DATA*
4…..*D*= $l_2$ ($P_1$, $p_i$)
5…..if *D* > *MAX* then
6………*MAX=D*
7………$P_2$= $p_i$
8. Stop=false
9. Do while (not stop)
10…$MAX_2$=0
11… for each point $p_i$ in *DATA*
12…..*D*= $l_2$ ($P_2$, $p_i$)
13…..if *D* > $MAX_2$ then
14………$MAX_2$=*D*
15………$P_3$= $p_i$
16….if ($MAX_2$> *MAX*) then
17…… $P_1$ = $P_2$
18…… $P_2$ = $P_3$
19…… *MAX* = $MAX_2$
20…else Stop=true
21.End while
22.Print *MAX*
---

*2.3. Tabu search implementation*

This algorithm is an attempt to increase the accuracy of The Hill climbing approach by storing all points with the same maximum distance. Hence, it is similar to the Algorithm (2) except for two differences: a) it uses a queue to store all points with the maximum similar distance, and b) it uses a binary array to keep track of visited points. See Algorithm 3. Obviously, this algorithm consumes more time, as the compared point might have two or more points with the same maximum distance, and this increases the number of iterations per point; this situation is more likely to occur in large datasets. The time complexity is similar to Equation (10) but with larger k in many cases, as it needs extra O(k.n) time to find all similar maximum distances, this makes the total time complexity

$$T(n,d) = O(k.n.d + k.n) \qquad \ldots (11)$$

Algorithm 3: Tabu search.
- - - - - - - - - - - - - - - - - - - - - - - - - - - - - - - - - - - - - - - - - - - - - - - - - - - - - - - - - - - - - - -
Input: DATA; A data set of *n* points and *d* dimensions of real numbers.
Output: MAX; The approximate diameter (the distance between the approximate furthest points)
- - - - - - - - - - - - - - - - - - - - - - - - - - - - - - - - - - - - - - - - - - - - - - - - - - - - - - - - - - - - - - -
1. $P_1$ = random point (1, n)
2. *MAX*=0
3. Initialize *Visited* Boolean array to be false for all points
4. Initialize *Distances* array to be 0 for all points
5. for each point $p_i$ in *DATA*
6….. $Distances_i = l_2 (P_1, p_i)$
7…..if $Distances_i > MAX$ then
8………$MAX = Distances_i$
9. for i=1 to n
10… if $Distances_i = MAX$ then
11………$P_2 = p_i$
12………Push ($P_2$) to *QUEUE*
13. Do while (*QUEUE* is not empty)
14…$P_3$= *QUEUE.pop()*
15… if (*Visited* [$P_3$]) go to step 13 else *Visited* [$P_3$] = true
16… Initialize *Distances* array to be 0 for all points
17… for each point $p_i$ in *DATA*
18…… if (*Visited* [$p_i$]) go to step 17
19…… $Distances_i = l_2 (P_3, p_i)$
20……if $Distances_i > MAX$ then $MAX = Distances_i$
21…for each *i* in *Distances* Push ($p_i$) to *QUEUE* iff $Distances_i = MAX$
22. End while
23. Print *MAX*
- - - - - - - - - - - - - - - - - - - - - - - - - - - - - - - - - - - - - - - - - - - - - - - - - - - - - - - - - - - - - - -

*2.4. Beam search implementation*

Doing multiple runs on different random points would enhance the performance of Algorithm 2. However, such an approach would not share information among runs, which allows for visiting the same point several times, and this increases time consumed without a significant increase to the accuracy. Therefore, we opt for Beam search approach, which seeds different random points to start "climbing" with. Such an approach can share information of visited points and save some time instead of re-calculating unneeded distances. Here we opt for 20 as the Beam size (B), this number can be changed to any other number. However, we think this number is relatively small, and the algorithm might provide better results. See Algorithm 4. The time complexity is not necessarily 20 times the time complexity of Algorithm 2 because of the visited points' information used by the algorithm; however, it should take longer. The time complexity can be defined by

$$T(n, d) = O(k.n.d + B.n.d) \qquad \ldots (12)$$

Algorithm 4: Beam search
- - - - - - - - - - - - - - - - - - - - - - - - - - - - - - - - - - - - - - - - - - - - - - - - - - - - - - - - - - - - - -
Input: DATA; A data set of *n* points and *d* dimensions of real numbers.
Output: MAX; The approximate diameter (the distance between the approximate furthest points).
- - - - - - - - - - - - - - - - - - - - - - - - - - - - - - - - - - - - - - - - - - - - - - - - - - - - - - - - - - - - - -
1. B=20
2. Initialize *Visited* Boolean array to be false for all points
3. for i=1 to B
4… $P_1$ = random point (1, n)
5.…*MAX*=0
6.…for each point $p_i$ in *DATA*
7.……..D= $l_2$ ($P_1$, $p_i$)
8.……..if *D > MAX* then *MAX=D;* $P_2$= $p_i$ ;
9.… Push ($P_2$) to *QUEUE*
10. Do while (*QUEUE* is not empty)
11…$P_3$= *QUEUE.pop()*
12… if (*Visited* [$P_3$]) go to step 10 **else** *Visited* [$P_3$] = true; flag=false
16… for each point $p_i$ in *DATA*
18…… if (*Visited* [$p_i$]) go to step 16
19…… D= $l_2$ ($P_3$, $p_i$)
20.……if *D > MAX* then *MAX= D,* $P_1$= $p_i$ ; flag=true;
21…if flag=true then Push ($P_1$) to *QUEUE*
22. End while
23. Print *MAX*
- - - - - - - - - - - - - - - - - - - - - - - - - - - - - - - - - - - - - - - - - - - - - - - - - - - - - - - - - - - - - -

## 3. Data

Since this work concerns finding the diameter of machine learning datasets, we used several real-world datasets downloaded from the UCI Machine Learning Repository (Lichman, 2013) to evaluate the implemented algorithms, in addition to the "Colors" and "Nasa" datasets, which are obtained from SISAP (Figueroa, Navarro, & Chavez, 2007). The chosen datasets are varied in dimensions, ranges and data types. Table 2 describes the data.

Table 2 Sample of machine learning datasets with different data types, n, d and ranges.

| #Dataset | Name | n | d | Data Type | Range |
|---|---|---|---|---|---|
| D1 | Iris | 150 | 4 | real | [0.1,7.9] |
| D2 | Haberman | 306 | 3 | digits | [0,83] |
| D3 | Glass | 214 | 9 | real | [0,75.41] |
| D4 | Liver | 345 | 6 | digits | [0,297] |
| D5 | Balance | 625 | 4 | digits | [1,5] |
| D6 | Wholesale | 440 | 7 | digits | [1,112151] |
| D7 | Vowel | 528 | 10 | real | [-5.211,5.074] |
| D8 | Banknote | 1372 | 4 | real | [-13.7731,17.9274] |
| D9 | Diabetes | 768 | 8 | real | [0,846] |
| D10 | Cancer | 683 | 9 | digits | [0,9] |
| D11 | Vote | 399 | 16 | digits | [0,2] |
| D12 | Heart | 270 | 25 | real | [0,564] |
| D13 | BCW | 699 | 10 | digits | [1,13454352] |
| D14 | Monkey1 | 556 | 17 | Binary | [0,1] |
| D15 | Ionosphere | 351 | 34 | real | [-1,1] |
| D16 | Sonar | 208 | 60 | real | [0,1] |
| D17 | Vehicle | 846 | 18 | digits | [0,1018] |
| D18 | German | 1000 | 24 | digits | [0,184] |
| D19 | Phoneme | 5404 | 5 | real | [-1.82,4.38] |
| D20 | Parkinson | 1040 | 27 | real | [0,1490] |
| D21 | Australian | 690 | 42 | real | [0,100001] |
| D22 | QSAR | 1055 | 41 | real | [-5.256,147] |
| D23 | Segmen | 2310 | 19 | real | [-49.68,1386.33] |
| D24 | Waveform21 | 5000 | 21 | real | [-4.2,9.06] |
| D25 | Waveform40 | 5000 | 40 | real | [-3.97,8.82] |
| D26 | EEG | 14980 | 14 | real | [86.67,72] |
| D27 | letter-recognition | 20000 | 16 | digits | [0,15] |
| D28 | Nasa | 40150 | 20 | real | [-1.33224,1.8424] |
| D29 | Colors | 112682 | 112 | real | [0,1] |

## 4. Results and discussion

We implemented the greedy algorithms using C++ language on a personal computer with Intel® Pentium® CPU G630 @ 2.70 GHz, 4 GB of RAM and a 32-bit operating system. The performance of these algorithms is evaluated on the datasets shown in Table 2. The evaluation is based on both accuracy and time consumed. We will refer to the greedy algorithms (1, 2, 3 and 4) as A1, A2, A3, and A4, respectively. Tables 3 and 4 show the performance of these algorithms in terms of time consumed and accuracy achieved.

Table 3 Time and number of iterations used by each of the proposed algorithms.

| Method | Time | Iterations |
|---|---|---|

| Data | BF | A1 | A2 | A3 | A4 | BF | A1 | A2 | A3 | A4 |
|---|---|---|---|---|---|---|---|---|---|---|
| D1 | 14 | 0 | 0 | 0 | 2 | 11175 | 2 | 4 | 4 | 25 |
| D2 | 55 | 0 | 0 | 0 | 5 | 46665 | 2 | 3 | 4 | 23 |
| D3 | 38 | 0 | 0 | 0 | 5 | 22791 | 2 | 3 | 4 | 24 |
| D4 | 85 | 0 | 1 | 0 | 7 | 59340 | 2 | 3 | 4 | 23 |
| D5 | 243 | 0 | 1 | 0 | 14 | 195000 | 2 | 3 | 4 | 31 |
| D6 | 144 | 0 | 1 | 0 | 10 | 96580 | 2 | 3 | 4 | 23 |
| D7 | 249 | 0 | 2 | 1 | 15 | 139128 | 2 | 3 | 4 | 23 |
| D8 | 1180 | 0 | 3 | 2 | 26 | 940506 | 2 | 3 | 4 | 25 |
| D9 | 473 | 0 | 2 | 1 | 20 | 294528 | 2 | 3 | 4 | 24 |
| D10 | 365 | 0 | 2 | 1 | 22 | 232903 | 2 | 3 | 4 | 30 |
| D11 | 161 | 0 | 2 | 1 | 15 | 79401 | 2 | 3 | 4 | 25 |
| D12 | 92 | 0 | 1 | 1 | 12 | 36315 | 2 | 3 | 4 | 23 |
| D13 | 399 | 0 | 2 | 1 | 19 | 243951 | 2 | 3 | 4 | 23 |
| D14 | 306 | 0 | 1 | 22 | 29 | 154290 | 2 | 2 | 29 | 36 |
| D15 | 239 | 0 | 3 | 2 | 30 | 61425 | 2 | 2 | 4 | 28 |
| D16 | 125 | 0 | 3 | 2 | 27 | 21528 | 2 | 3 | 4 | 27 |
| D17 | 859 | 0 | 4 | 3 | 36 | 357435 | 2 | 3 | 4 | 23 |
| D18 | 1244 | 0 | 6 | 3 | 48 | 499500 | 2 | 3 | 4 | 25 |
| D19 | 19862 | 1 | 14 | 9 | 134 | 14598906 | 2 | 3 | 4 | 29 |
| D20 | 1685 | 0 | 7 | 5 | 59 | 540280 | 2 | 3 | 4 | 23 |
| D21 | 789 | 1 | 6 | 5 | 42 | 237705 | 2 | 3 | 4 | 23 |
| D22 | 2122 | 1 | 10 | 6 | 75 | 555985 | 2 | 3 | 4 | 23 |
| D23 | 6327 | 1 | 12 | 16 | 95 | 2666895 | 2 | 3 | 6 | 23 |
| D24 | 45045 | 3 | 48 | 24 | 414 | 12497500 | 2 | 3 | 4 | 26 |
| D25 | 86034 | 5 | 108 | 90 | 889 | 12497500 | 2 | 4 | 4 | 31 |
| D26 | 290055 | 6 | 69 | 49 | 520 | 112192710 | 2 | 3 | 4 | 23 |
| D27 | 468757 | 7 | 103 | 70 | 935 | 199990000 | 2 | 3 | 4 | 28 |
| D28 | 2176035 | 16 | 351 | 190 | 2215 | 805991175 | 2 | 4 | 4 | 27 |
| D29 | 49979645 | 175 | 2281 | 133296 | 23015 | 2053592925 | 2 | 3 | 182 | 31 |

As can be seen from Table 3, A1 consumed the least time, and this is expected since the dominant O(n.d) time is used only twice. The Beam Search algorithm (A4) consumed more time and iterations, because it starts with 20 random points; however, interestingly, the number of iterations is not proportional with the initial seed (20) because of the memory used, as visited points are not re-calculated by A4. A2 and A3 came in between in terms of time and number of iterations, and were almost similar, with some increase in number of iterations, and this is due to the maximum equal distances, which depend mainly on the dataset itself. There is a large increase in number of iterations (and therefore time) when calculating the diameter for D4 (Binary data) and D29 (data rang [0,1]), such kind of data with many zeros and ones allows for similar distances between points, as the permutations of the same number of zeros and ones make different points in the feature space but with similar Euclidean distance.

Table 4 The diameters of the tested datasets after applying the proposed methods.

| Method | Diameter | Approximation |
|---|---|---|

| Data | BF | A1 | A2 | A3 | A4 | BF | A1 | A2 | A3 | A4 |
|---|---|---|---|---|---|---|---|---|---|---|
| D1 | 7.09E+00 | 7.09E+00 | 7.09E+00 | 7.09E+00 | 7.09E+00 | 0 | 0.0 | 0.0 | 0.0 | 0.0 |
| D2 | 6.40E+01 | 6.40E+01 | 6.40E+01 | 6.40E+01 | 6.40E+01 | 0 | 0.0 | 0.0 | 0.0 | 0.0 |
| D3 | 1.20E+01 | 1.20E+01 | 1.20E+01 | 1.20E+01 | 1.20E+01 | 0 | 0.0 | 0.0 | 0.0 | 0.0 |
| D4 | 2.95E+02 | 2.92E+02 | 2.95E+02 | 2.95E+02 | 2.95E+02 | 0 | 3.9 | 0.0 | 0.0 | 0.0 |
| D5 | 8.00E+00 | 8.00E+00 | 8.00E+00 | 8.00E+00 | 8.00E+00 | 0 | 0.0 | 0.0 | 0.0 | 0.0 |
| D6 | 1.29E+05 | 1.29E+05 | 1.29E+05 | 1.29E+05 | 1.29E+05 | 0 | 0.0 | 0.0 | 0.0 | 0.0 |
| D7 | 9.35E+00 | 9.35E+00 | 9.35E+00 | 9.35E+00 | 9.35E+00 | 0 | 0.0 | 0.0 | 0.0 | 0.0 |
| D8 | 3.45E+01 | 2.61E+01 | 3.45E+01 | 3.45E+01 | 3.45E+01 | 0 | 8.4 | 0.0 | 0.0 | 0.0 |
| D9 | 8.68E+02 | 8.68E+02 | 8.68E+02 | 8.68E+02 | 8.68E+02 | 0 | 0.0 | 0.0 | 0.0 | 0.0 |
| D10 | 2.57E+01 | 2.56E+01 | 2.56E+01 | 2.56E+01 | 2.57E+01 | 0 | 0.1 | 0.1 | 0.1 | 0.0 |
| D11 | 7.21E+00 | 7.21E+00 | 7.21E+00 | 6.40E+00 | 7.21E+00 | 0 | 0.0 | 0.0 | 0.8 | 0.0 |
| D12 | 4.40E+02 | 4.17E+02 | 4.40E+02 | 4.40E+02 | 4.40E+02 | 0 | 22.9 | 0.0 | 0.0 | 0.0 |
| D13 | 1.34E+07 | 1.34E+07 | 1.34E+07 | 1.34E+07 | 1.34E+07 | 0 | 0.0 | 0.0 | 0.0 | 0.0 |
| D14 | 3.46E+00 | 2.00E+00 | 3.46E+00 | 3.46E+00 | 3.46E+00 | 0 | 1.5 | 0.0 | 0.0 | 0.0 |
| D15 | 9.75E+00 | 8.81E+00 | 9.75E+00 | 9.75E+00 | 9.75E+00 | 0 | 0.9 | 0.0 | 0.0 | 0.0 |
| D16 | 3.53E+00 | 2.81E+00 | 3.53E+00 | 3.06E+00 | 3.53E+00 | 0 | 0.7 | 0.0 | 0.5 | 0.0 |
| D17 | 8.80E+02 | 8.80E+02 | 8.80E+02 | 8.80E+02 | 8.80E+02 | 0 | 0.0 | 0.0 | 0.0 | 0.0 |
| D18 | 1.87E+02 | 1.86E+02 | 1.87E+02 | 1.87E+02 | 1.87E+02 | 0 | 1.2 | 0.0 | 0.1 | 0.0 |
| D19 | 6.17E+00 | 5.49E+00 | 5.69E+00 | 5.97E+00 | 6.17E+00 | 0 | 0.7 | 0.5 | 0.2 | 0.0 |
| D20 | 2.13E+03 | 2.10E+03 | 2.13E+03 | 2.13E+03 | 2.13E+03 | 0 | 34.3 | 0.0 | 0.0 | 0.0 |
| D21 | 1.00E+05 | 1.00E+05 | 1.00E+05 | 1.00E+05 | 1.00E+05 | 0 | 9.0 | 0.0 | 0.0 | 0.0 |
| D22 | 1.73E+02 | 1.73E+02 | 1.73E+02 | 1.73E+02 | 1.73E+02 | 0 | 0.0 | 0.0 | 0.0 | 0.0 |
| D23 | 1.52E+03 | 1.51E+03 | 1.52E+03 | 1.52E+03 | 1.52E+03 | 0 | 14.8 | 0.0 | 0.0 | 0.0 |
| D24 | 2.30E+01 | 2.18E+01 | 2.30E+01 | 2.30E+01 | 2.30E+01 | 0 | 1.3 | 0.0 | 0.0 | 0.0 |
| D25 | 2.38E+01 | 2.09E+01 | 2.38E+01 | 2.38E+01 | 2.38E+01 | 0 | 2.9 | 0.0 | 0.0 | 0.0 |
| D26 | 1.10E+06 | 1.10E+06 | 1.10E+06 | 1.10E+06 | 1.10E+06 | 0 | 0.0 | 0.0 | 0.0 | 0.0 |
| D27 | 3.34E+01 | 3.31E+01 | 3.34E+01 | 3.34E+01 | 3.34E+01 | 0 | 0.3 | 0.0 | 0.0 | 0.0 |
| D28 | 2.83E+00 | 2.55E+00 | 2.83E+00 | 2.83E+00 | 2.83E+00 | 0 | 0.3 | 0.0 | 0.0 | 0.0 |
| D29 | 1.41E+00 | 1.41E+00 | 1.41E+00 | 1.41E+00 | 1.41E+00 | 0 | 0.0 | 0.0 | 0.0 | 0.0 |

The approximation in Table 4 is calculated by subtracting the output diameter of each algorithm from the actual diameter presented by the BF; the smaller the approximation, the larger the accuracy of an algorithm will be.

Interestingly, Algorithms A2 and A3 found the exact diameter for most datasets tested, and even for those datasets for which the diameter was not determined accurately, the approximations were very low (in the range [0.1, 0.5] for A2, and [0.1, 0.8] for A3. It is worth noting that A4 achieved the optimal diameter for all the datasets tested; however, this does not prove that A4 is an exact algorithm, since it uses random seeding; there is no guarantee to find an optimal solution all the time. We tested A4 on another synthesized data (n=1198, d=14, range [-10, 10] real values), and it was not exact (the approximation was very small = 0.68). Therefore, there is no need to prove otherwise.

A closer look at the results in Tables 3 and 4 reveals that the faster algorithms are less accurate and that the slower ones are more accurate. This prevents us from recommending an algorithm for finding the diameters of machine learning datasets, which are required for other machine learning tasks. Our concern in this work is two-fold: accuracy achieved and time consumed. Therefore, to further evaluate the algorithms, we opt for algorithm's efficiency, which is based on the algorithm's accuracy and time consumption, so the accuracy of an algorithm can be defined by

$$Accuracy = 1 - \frac{approximation}{actual\ diameter} \qquad \ldots (13)$$

where approximation = the actual diameter – the output diameter, the actual diameter is the ground truth diameter, which is found by an exact algorithm such as the BF.

The efficiency of an algorithm can be defined by

$$Efficiency = \frac{Accuracy}{1+T(A)/T(BF)} \qquad \ldots (14)$$

where T(A) is the time consumed by an Algorithm A, and T(BF) is the actual time needed by the BF to find the optimal solution under the same circumstances and using the same resources. Accuracy and efficiency for the previous algorithms are calculated in Table 5.

Table 5 Accuracy and efficiency of the implemented algorithms.

| Method Data | Accuracy | | | | | Efficiency | | | | |
|---|---|---|---|---|---|---|---|---|---|---|
| | BF | A1 | A2 | A3 | A4 | BF | A1 | A2 | A3 | A4 |
| D1 | 1.000 | 1.000 | 1.000 | 1.000 | 1.000 | 0.500 | 1.000 | 1.000 | 1.000 | 0.875 |
| D2 | 1.000 | 1.000 | 1.000 | 1.000 | 1.000 | 0.500 | 1.000 | 1.000 | 1.000 | 0.917 |
| D3 | 1.000 | 1.000 | 1.000 | 1.000 | 1.000 | 0.500 | 1.000 | 1.000 | 1.000 | 0.884 |
| D4 | 1.000 | 0.987 | 1.000 | 1.000 | 1.000 | 0.500 | 0.987 | 0.988 | 1.000 | 0.924 |
| D5 | 1.000 | 1.000 | 1.000 | 1.000 | 1.000 | 0.500 | 1.000 | 0.996 | 1.000 | 0.946 |
| D6 | 1.000 | 1.000 | 1.000 | 1.000 | 1.000 | 0.500 | 1.000 | 0.993 | 1.000 | 0.935 |
| D7 | 1.000 | 1.000 | 1.000 | 1.000 | 1.000 | 0.500 | 1.000 | 0.992 | 0.996 | 0.943 |
| D8 | 1.000 | 0.755 | 1.000 | 1.000 | 1.000 | 0.500 | 0.755 | 0.997 | 0.998 | 0.978 |
| D9 | 1.000 | 1.000 | 1.000 | 1.000 | 1.000 | 0.500 | 1.000 | 0.996 | 0.998 | 0.959 |
| D10 | 1.000 | 0.995 | 0.995 | 0.995 | 1.000 | 0.500 | 0.995 | 0.990 | 0.993 | 0.943 |
| D11 | 1.000 | 1.000 | 1.000 | 0.888 | 1.000 | 0.500 | 1.000 | 0.988 | 0.882 | 0.915 |
| D12 | 1.000 | 0.948 | 1.000 | 1.000 | 1.000 | 0.500 | 0.948 | 0.989 | 0.989 | 0.885 |
| D13 | 1.000 | 1.000 | 1.000 | 1.000 | 1.000 | 0.500 | 1.000 | 0.995 | 0.998 | 0.955 |
| D14 | 1.000 | 0.577 | 1.000 | 1.000 | 1.000 | 0.500 | 0.577 | 0.997 | 0.933 | 0.913 |
| D15 | 1.000 | 0.904 | 1.000 | 1.000 | 1.000 | 0.500 | 0.904 | 0.988 | 0.992 | 0.888 |
| D16 | 1.000 | 0.795 | 1.000 | 0.866 | 1.000 | 0.500 | 0.795 | 0.977 | 0.852 | 0.822 |
| D17 | 1.000 | 1.000 | 1.000 | 1.000 | 1.000 | 0.500 | 1.000 | 0.995 | 0.997 | 0.960 |
| D18 | 1.000 | 0.994 | 1.000 | 0.999 | 1.000 | 0.500 | 0.994 | 0.995 | 0.997 | 0.963 |
| D19 | 1.000 | 0.890 | 0.923 | 0.967 | 1.000 | 0.500 | 0.890 | 0.922 | 0.967 | 0.993 |
| D20 | 1.000 | 0.984 | 1.000 | 1.000 | 1.000 | 0.500 | 0.984 | 0.996 | 0.997 | 0.966 |
| D21 | 1.000 | 1.000 | 1.000 | 1.000 | 1.000 | 0.500 | 0.999 | 0.992 | 0.994 | 0.949 |
| D22 | 1.000 | 1.000 | 1.000 | 1.000 | 1.000 | 0.500 | 1.000 | 0.995 | 0.997 | 0.966 |
| D23 | 1.000 | 0.990 | 1.000 | 1.000 | 1.000 | 0.500 | 0.990 | 0.998 | 0.997 | 0.985 |
| D24 | 1.000 | 0.945 | 1.000 | 1.000 | 1.000 | 0.500 | 0.945 | 0.999 | 0.999 | 0.991 |
| D25 | 1.000 | 0.877 | 1.000 | 1.000 | 1.000 | 0.500 | 0.877 | 0.999 | 0.999 | 0.990 |
| D26 | 1.000 | 1.000 | 1.000 | 1.000 | 1.000 | 0.500 | 1.000 | 1.000 | 1.000 | 0.998 |
| D27 | 1.000 | 0.990 | 1.000 | 1.000 | 1.000 | 0.500 | 0.990 | 1.000 | 1.000 | 0.998 |
| D28 | 1.000 | 0.901 | 1.000 | 0.999 | 1.000 | 0.500 | 0.901 | 1.000 | 0.999 | 0.999 |
| D29 | 1.000 | 1.000 | 1.000 | 1.000 | 1.000 | 0.500 | 1.000 | 1.000 | 0.997 | 1.000 |
| Average | 1.000 | 0.945 | 0.993 | 0.984 | 1.000 | 0.500 | 0.945 | **0.988** | 0.979 | 0.946 |

As can be seen in Table 5, the most accurate Algorithm (A4) is not necessarily the most efficient, since it consumes more time, neither the BF for the same reason. At the same time, the faster algorithm (A1) is not necessarily the most efficient as it is the least accurate. This leaves two options (A2 and A3), particularly A2 (the Hill climbing search), which achieved the highest efficiency. A2 is slightly more efficient than A3 because of considering only points with larger distances, while A3 was considering all points with maximum equal points. This sometimes leads to better approximation, but such insignificant enhancement does not justify the extra time consumed comparing to that of the A2.

Since most of the algorithms found in the literature proposed in theory, have no available practical code to be used for comparison and are difficult to be implemented (Har-Peled, 2001), we opt for using the reported time complexities on the sizes and dimensions of the machine learning datasets. Having known that, calculating the number of operations that are used by an algorithm might approximate the real running time to some degree of accuracy. For conciseness, we choose datasets with (small n, small d), (small n, large d), (large n, small d) and (large n, large d). Since ε is in the range (0, 1] for most of the methods found in the literature, we opt for ε =0.5 as a midpoint to calculate the number of operations for each method that uses such approximation. Table 6 presents the calculations.

Table 6 Algorithms comparison in terms of number of operations on some datasets.

| Dataset<br>Method | D1 | D16 | D19 | D25 | D27 | D28 | D29 |
|---|---|---|---|---|---|---|---|
| BF | 4.47E+04 | 1.29E+06 | 7.30E+07 | 5.00E+08 | 3.20E+09 | 1.61E+10 | 7.11E+11 |
| M1 | **1.54E+02** | 2.88E+17 | **5.41E+03** | 2.75E+11 | **3.64E+04** | **3.02E+05** | 1.30E+33 |
| M2 | 3.07E+03 | 1.22E+12 | 2.68E+05 | 4.56E+10 | 5.17E+07 | 4.45E+08 | 9.64E+22 |
| M3 | 1.31E+05 | 3.33E+05 | 3.04E+08 | 3.07E+08 | 5.72E+09 | 2.47E+10 | 2.13E+11 |
| M4 | 1.13E+04 | 2.16E+04 | 1.46E+07 | 1.25E+07 | 2.00E+08 | 8.06E+08 | 6.35E+09 |
| M5 | 2.68E+04 | 1.39E+05 | 2.95E+07 | 2.75E+07 | 4.05E+08 | 1.62E+09 | 1.29E+10 |
| M6 | 1.56E+02 | 4.08E+17 | 5.42E+03 | 3.89E+11 | 4.32E+04 | 4.11E+05 | 1.84E+33 |
| M7 | 1.73E+02 | 4.38E+26 | 5.47E+03 | 4.08E+17 | 5.95E+06 | 3.80E+08 | 1.32E+50 |
| M8 | 1.73E+02 | 4.38E+26 | 5.47E+03 | 4.08E+17 | 5.95E+06 | 3.80E+08 | 1.32E+50 |
| A1 | 2.98E+03 | **1.92E+04** | 1.62E+05 | **3.29E+05** | 8.95E+05 | 2.04E+06 | **1.64E+07** |
| A2 | 2.40E+03 | 3.74E+04 | 8.11E+04 | 8.00E+05 | 9.60E+05 | 3.21E+06 | 3.79E+07 |
| A3 | 2.40E+03 | 4.99E+04 | 1.08E+05 | 8.00E+05 | 1.28E+06 | 3.21E+06 | 2.30E+09 |
| A4 | 1.50E+04 | 3.37E+05 | 7.84E+05 | 6.20E+06 | 8.96E+06 | 2.17E+07 | 3.91E+08 |

\* Calculations for methods M1 to M8 are made using equations (1-8) from Table 1.
\* Calculations of BF and A1-A4 are based on actual number of operations made by each algorithm from the implementation.

As can be seen from Table 6, while M1 (Imanparast, Hashemi, & Mohades, 2016) performs the best on 4 datasets (D1, D19, D27 and D28, with d =4, 5, 16 and 20 respectively), it performs very badly on the other datasets. This is due to the effect of the value of d on this algorithm, as this algorithm is exponential in terms of d, and is not affected much by the value of n (See Equation 1). This phenomenon can be noticed with all methods found in the literature (M1-M8), particularly at columns D16, D25 and D29, which have d= 60, 40 and 112, respectively. Such exponential behavior makes these algorithms perform much less than the exact BF, which is not acceptable in application. However, the implemented algorithms are affected by both n and d linearly, and their performances were consistent regarding the different dimensions. This behavior is expected and complies with time complexity analysis of A1-A4. See Equations 9, 10, 11 and 12.

The number of operations presented in Table 6 does not show the full picture. For example, A1 performs better than A2, A3 and A4. This is because it is faster, but due to its lower accuracy, its efficiency becomes lower, too. To compare the efficiency of the other algorithms with the implemented ones, we assume that the approximation is $\varepsilon = 0.5$, knowing that most of the algorithms (M1 to M8) uses $1+\varepsilon$ or more. The accuracy of each algorithm is calculated using Equation 13, and the efficiency is calculated using Equation 14. Table 7 presents the calculations.

Table 7 Efficiency of different methods

| Dataset<br>Method | D1 | D16 | D19 | D25 | D27 | D28 | D29 | Avg. |
|---|---|---|---|---|---|---|---|---|
| BF | 0.500 | 0.500 | 0.500 | 0.500 | 0.500 | 0.500 | 0.500 | 0.500 |
| M1 | 0.926 | 0.000 | 0.919 | 0.002 | 0.985 | 0.823 | 0.000 | 0.522 |
| M2 | 0.870 | 0.000 | 0.916 | 0.011 | 0.969 | 0.801 | 0.000 | 0.509 |
| M3 | 0.237 | 0.682 | 0.178 | 0.606 | 0.354 | 0.325 | 0.497 | 0.411 |
| M4 | 0.743 | 0.844 | 0.766 | 0.955 | 0.927 | 0.784 | 0.641 | 0.808 |
| M5 | 0.581 | 0.775 | 0.654 | 0.928 | 0.874 | 0.748 | 0.635 | 0.742 |
| M6 | 0.926 | 0.000 | 0.919 | 0.001 | 0.985 | 0.823 | 0.000 | 0.522 |
| M7 | 0.926 | 0.000 | 0.919 | 0.000 | 0.983 | 0.804 | 0.000 | 0.519 |
| M8 | 0.926 | 0.000 | 0.919 | 0.000 | 0.983 | 0.804 | 0.000 | 0.519 |
| A1 | 0.938 | 0.783 | 0.888 | 0.877 | 0.990 | 0.901 | **1.000** | 0.911 |
| A2 | **0.949** | **0.972** | 0.922 | **0.998** | **1.000** | **1.000** | **1.000** | **0.977** |
| A3 | **0.949** | 0.834 | 0.966 | **0.998** | **1.000** | 0.999 | 0.997 | 0.963 |
| A4 | 0.749 | 0.793 | **0.989** | 0.988 | 0.997 | 0.999 | 0.999 | 0.931 |

As noted in Table 7, M1 is still efficient on the same datasets with relatively smaller dimensions, but not more efficient than (A1-A4). This is due to the accuracies of these algorithms, which reach to 1 in most cases, while it depends on the fixed approximation 0.5 for M1. The same applies to the other methods (M2-M8), while their efficiencies on datasets with larger dimensions are very low, and sometimes less than 0.5 the ideal efficiency of the BF, or even zeros with large dimensions (D16, D25 and D29) – i.e., less efficient than the impractical algorithm BF. On average, we can see that A2 is the most efficient of them all, but the efficiencies of the rest of the implemented algorithms are not different significantly, and this makes A1, A2, A3 and A4 appropriate for AI applications.

It is worth mentioning that the efficiencies of A1, A2, A3 and A4 calculated in Table 7 are slightly different from those presented in Table 5, because their accuracies and efficiencies are calculated based on number of operations and not based on real-time consumed by the CPU. It is also worth mentioning that if M1-M8 are coded and implemented on the same datasets, the results might be changed, but not significantly, as the reported theoretical time complexity tills.

## 5. Conclusion

AI and machine learning databases normally have relatively small n around 1000-5000 (the median size of 365 UCI machine learning datasets is 1540). This number represents the examples of the sample that are meant to represent the population of a real-world problem. Obtaining such learning examples is costly and therefore their quantity tends to be relatively small. However, these kinds of databases have relatively large number of dimensions, around 20-50 (the median dimensions of 365 UCI machine learning datasets is 21), and sometimes much larger. This is due to the features needed for training, such as CBIR, speech recognition, iris code, etc., so finding the diameter using algorithms that do not scale well to large dimensions of the data makes such algorithms impractical, particularly for AI and machine learning applications.

Most of the algorithms found in the literature do not scale well with large values of the dimension of the data, and the time complexity grows exponentially in most cases. Therefore, we implemented 4 simple approximate algorithms to be used for finding the diameter of a dataset in any dimension. The time complexity analysis of the implemented algorithms confirms a near-linear time for them all, and the algorithms scale near-linearly with number of points and dimensions.

We find through experiments that the implemented algorithms approximate well, fast and therefore efficient. We also find that the four implemented algorithms vary in speed and accuracy; nevertheless, they all of near-linear time and having high accuracy on average. A1 is the fastest with the lowest accuracy, and A4 is the slowest with a highest accuracy. A2 and A3 are in the middle, A2 is the most efficient taking into consideration both speed and accuracy, and therefore, we recommend it to find the diameter for different applications.

We also compared the implemented algorithms with some of the most common state-of-the-art algorithms in terms of number of operations and their approximations. We found that the implemented algorithms are more efficient than those, particularly, when applied on different machine learning datasets.

Without implementing the state-of-the-art algorithms (in Table 1), the comparisons discussed in this work might be inadequate to a certain degree. This major limitation will be addressed in the future work regarding the difficulties of implementing such theoretical algorithms. This work is part of a larger project about approximate nearest neighbor search. One of the implemented methods will be used with some other techniques to provide a new solution for this problem; we tend to use A2 (because of its high efficiency) for our future project.

## Acknowledgment

The author would like to thank Dr. Hubert Anysz, with Warsaw University of Technology, for his valuable discussions and inspiration of the implemented methods.

## References

Agarwal, P. K., Har-Peled, S., & Varadarajan, K. R. (2005). Geometric approximation via coresets. *Combinatorial and computational geometry, 52*, 1-30.

Agarwal, P. K., Matoušek, J., & Suri, S. (1992). Farthest neighbors, maximum spanning trees and related problems in higher dimensions. *Computational Geometry, 1*(4), 189-201.

Aggarwal, C. C. (2016). *Outlier Analysis* (Second Edition ed.). New York: Springer.

Bespamyatnikh, S. (1998). An efficient algorithm for the three-dimensional diameter problem. *9th Symp. on Discrete Algorithms*, (pp. 137–146).

Broder, A. Z., Glassman, S. C., Manasse, M. S., & Zweig, G. (1997). Syntactic clustering of the web. *Computer Networks and ISDN Systems, 29*(8), 1157-1166.

Chan, T. M. (2002). Approximating the diameter, width, smallest enclosing cylinder, and minimum-width annulus. *International journal of computational geometry and applications, 12*(1), 67-85.

Chazelle, B., Edelsbrunner, H., Guibas, L., & Sharir, M. (1992). Diameter, width, closest line pair, and parametric searching. *eighth annual symposium on Computational geometry* (pp. 120-129). ACM.

Clarkson, K. L., & Shor, P. W. (1989). Applications of random sampling in computational geometry, II. *Discrete & Computational Geometry, 4*(5), 387-421.

Eğecioğlu, Ö., & Kalantari, B. (1989). Approximating the diameter of a set of points in the Euclidean space. *Information Processing Letters, 32*(4), 205-211.

Figueroa, K., Navarro, G., & Chavez, E. (2007). *Metric Spaces Library*. Retrieved from http://www.sisap.org/Metric\_Space\_Library.html

Finocchiaro, D. V., & Pellegrini, M. (2002). On computing the diameter of a point set in high dimensional Euclidean space. *Theoretical Computer Science, 287*, 501–514.

Gudivada, V. N., & Raghavan, V. V. (1995). Content based image retrieval systems. *Computer, 28*(9), 18-22.

Har-Peled, S. (2001). A practical approach for computing the diameter of a point set. *The seventeenth annual symposium on Computational geometry* (pp. 177-186). ACM.

Hassanat, A. B., & Tarawneh, A. S. (2016). Fusion of Color and Statistic Features for Enhancing Content-Based Image Retrieval Systems. *Journal of Theoretical & Applied Information Technology, 88*(3).

Imanparast, M., Hashemi, S. N., & Mohades, A. (2016). An efficient approximation for point set diameter in fixed dimensions. *preprint arXiv:1610.08543 [cs.CG]*. Retrieved from https://arxiv.org/abs/1610.08543

Le Bourdais, F. (2015, Jul 16). *The Farthest Neighbors Algorithm*. Retrieved Feb 1, 2018, from Frolian's blog: https://flothesof.github.io/farthest-neighbors.html

Lichman, M. (2013). (University of California, Irvine, School of Information and Computer Sciences) Retrieved from UCI Machine Learning Repository: http://archive.ics.uci.edu/ml

Matoušek, J., & Schwarzkopf, O. (1993). A deterministic algorithm for the three-dimensional diameter problem. *the twenty-fifth annual ACM symposium on Theory of computing* (pp. 478-484). ACM.

Pagh, R., Silvestri, F., Sivertsen, J., & Skala, M. (2015). Approximate furthest neighbor in high dimensions. *International Conference on Similarity Search and Applications* (pp. 3-14). Springer.

Pagh, R., Silvestri, F., Sivertsen, J., & Skala, M. (2017). Approximate furthest neighbor with application to annulus query. *Information Systems, 64*, 152-162.

Preparata, F., & Shamos, M. (1985). *Computational Geometry: an Introduction.* New York: Springer.

Ramos, E. (1997). Construction of 1-d lower envelopes and applications. *13th Symp. on Computational Geometry*, (pp. 57–66).

Ramos, E. A. (2000). Deterministic algorithms for 3-D diameter and some 2-D lower envelopes. *The 16th annual symposium on Computational Geometry* (pp. 290-299). ACM.

Supowit, K. J. (1990). New techniques for some dynamic closest-point and farthest-point problems. *The first annual ACM-SIAM symposium on Discrete algorithms* (pp. 84-90). Society for Industrial and Applied Mathematics.

Williams, R. (2018). On the Difference Between Closest, Furthest, and Orthogonal Pairs: Nearly-Linear vs Barely-Subquadratic Complexity. *Twenty-Ninth Annual ACM-SIAM Symposium on Discrete Algorithms* (pp. 1207-1215). Twenty-Ninth Annual ACM-SIAM Symposium on Discrete Algorithms.

Yao, A. C. (1982). On constructing minimum spanning trees in k-dimensional spaces and related problems. *SIAM Journal on Computing, 11*(4), 721-736.